\documentclass[sigconf,nonacm]{acmart}

\usepackage{booktabs}
\usepackage{xcolor}
\usepackage{amsmath}
\usepackage{multirow}

\newif\ifshowtodos
\showtodosfalse   
\ifshowtodos

\else

\fi

\begin{document}

\title[DealsFlow: Evidence-Gated Agentic Triage for Venture Capital]{DealsFlow: An Evidence-Gated Agentic Harness for Venture Capital Deal Triage}

\author{Ming Ze Tang}
\affiliation{%
  \institution{Queen's University Belfast}
  \city{Belfast}
  \country{United Kingdom}}
\email{m.tang@qub.ac.uk}

\author{David Cutting}
\affiliation{%
  \institution{Queen's University Belfast}
  \city{Belfast}
  \country{United Kingdom}}
\email{d.cutting@qub.ac.uk}

\author{Boyd Carson}
\affiliation{%
  \institution{Sapphire Capital Partners}
  \city{Belfast}
  \country{United Kingdom}}
\email{boyd@sapphirecapitalpartners.co.uk}

\author{Vasiliki Carson}
\affiliation{%
  \institution{Sapphire Capital Partners}
  \city{Belfast}
  \country{United Kingdom}}
\email{vasiliki@sapphirecapitalpartners.co.uk}

\renewcommand{\shortauthors}{Tang et al.}

\begin{abstract}
Large language models (LLMs) can read unstructured startup pitch decks, but letting them emit investment scores directly invites score inflation from promotional language, run-to-run variance, and rationales that are hard to audit. We describe DealsFlow, a human-in-the-loop agentic harness deployed at a UK early-stage fund, built on one design rule: the LLM reads the deck, cites evidence and rates each category, and code combines the ratings, caps them and only ever lowers them. Three mechanisms implement the rule: (i) a stage-aware corroboration ceiling that caps factual-category scores at 78, 70 or 60 out of 100 (early-stage, unknown-stage, growth-stage) unless a claim is matched to an external source; (ii) an asymmetric sceptic agent that may lower but never raise a category score; and (iii) a declared-dependency pre-fetch pattern in which every data source is retrieved and given a numbered index before a single generation pass, with the source list and the scores before and after each guardrail stored for audit. We evaluate the harness on a portfolio of $N=243$ SEIS/EIS investments with recorded outcomes. Outcomes in such data are censored by time: none of the 73 investments made in the last two years has been written off, so entry year alone separates write-offs with an AUC of 0.72. We therefore evaluate on the 142 investments at least three years old, where entry year showed no clear discrimination (AUC 0.45) and a cross-validated logistic regression on intake fields reaches an AUC of 0.64. Because corroboration is exactly the signal that hindsight can contaminate (successful companies leave more public evidence), we make leakage the centre of the evaluation. Our retrieval is not restricted to the decision date, so we treat leakage as a risk to be measured rather than one we have removed: we run a leakage audit that reports corroboration rates by outcome class and compares against a corroboration-only baseline, and we list the leakage channels we do not control. On these mature investments, the full pipeline's composite score has a Spearman correlation of $\rho=0.45$ (95\% CI [0.32, 0.57]) with the total-value multiple and an AUC of 0.71 [0.62, 0.79] for identifying write-offs, against $\rho=0.38$ and AUC 0.68 for an unconstrained LLM. The improvement is small but its paired interval excludes zero; it comes mostly from the second, sceptic call (we did not test whether its lower-only rule matters or whether any second review would do as well), while the ceiling alone adds little. A corroboration-only baseline reaches $\rho=0.22$. However, investments that later exceeded $2\times$ had 64\% of their factual categories corroborated, against 38\% for write-offs (omnibus Kruskal--Wallis test across the five outcome groups, $p=0.031$), and our retrieval is not date-restricted, so these figures may overstate screening skill. In a separate study of 20 inbound applications, total analyst time per application, including verification and correction, fell from 44.7 to 18.9 minutes, and two reviewers found 84\% of 240 cited claims fully supported by their sources, but 70\% of reports still needed at least one correction.
\end{abstract}


\maketitle

\section{Introduction}

UK early-stage funds investing under the Seed Enterprise Investment Scheme (SEIS) and Enterprise Investment Scheme (EIS) \cite{hmrcvcm} receive far more unsolicited applications than analysts can read closely. Top-of-funnel triage (reading the deck, checking the company register, searching for the founders, competitors and market) is repetitive and time-consuming, and it is where the founder's own optimistic framing has the most influence.

LLMs are an obvious fit for reading decks \cite{naveed2023overview}, but three of their known behaviours make them a poor fit for producing the final score unchecked. They tend to agree with the framing they are given \cite{sharma2024sycophancy,perez2023discovering,natan2026sycophant}; their outputs vary across runs even under nominally deterministic settings \cite{atil2025nondeterminism,he2025nondeterminism}; and their stated rationales need not reflect what drove the output \cite{turpin2023unfaithful}. In a firm that must be able to explain later why a deal was progressed or rejected, a single opaque LLM score is hard to defend \cite{rudin2019stop}.

DealsFlow is a decision-support harness built around one rule: \emph{the LLM reads and rates; code combines, caps and only ever lowers the ratings}. It never makes an investment decision; approvals are held for explicit human sign-off. This paper describes the harness and, more importantly, asks whether its scores carry information about real outcomes once the obvious ways an evaluation of this kind can fool itself are accounted for.

The most important of these is look-ahead bias. Our main guardrail rewards claims that can be corroborated externally. Companies that later succeed accumulate press coverage, filings and web presence; companies that fail often disappear. A backtest that retrieves today's web therefore risks scoring the outcome rather than the application. Our retrieval is not restricted to the decision date, so this risk applies to our own experiments; we measure it rather than assume it away. Recent work in financial LLM evaluation identifies exactly this channel, together with parametric memory in model weights and survivorship in the evaluation universe, as the most common threats to validity \cite{kong2026bias,lopezlira2025memorization,glasserman2023lookahead,sarkar2024lookahead}. We structure our evaluation around those threats.

\paragraph{Contributions.}
\begin{enumerate}
  \item \textbf{A design pattern for evidence-gated scoring} (\S\ref{sec:design}): a division of labour in which the LLM reads and rates while code combines the ratings, a stage-aware corroboration ceiling, a second-pass sceptic whose adjustments can only lower scores, and a declared-dependency pre-fetch that assembles the evidence before generation and stores the source list for audit. We state the arithmetic consequences of the ceiling explicitly (Proposition~\ref{prop:ceiling}).
  \item \textbf{A leakage audit on a recorded-outcome VC portfolio} (\S\ref{sec:evaldesign}): an explicit account of the leakage channels our retrieval does not control, corroboration rates by outcome class, a corroboration-only baseline that isolates how much of the score is explained by evidence availability alone, and a maturity restriction that reduces the influence of very recent investments, whose outcomes are still unresolved (\S\ref{sec:maturity}).
  \item \textbf{Empirical results on 243 holdings, with 142 mature holdings as the primary set} (\S\ref{sec:results}): rank correlation and AUC with bootstrap intervals, comparison with structured-field and corroboration-only baselines, an ablation of each guardrail, and an analyst assessment of total analyst time and factual-support quality (\S\ref{sec:time}, \S\ref{sec:support}).
\end{enumerate}

\paragraph{Scope of the claims.} We claim that the pattern is implementable in a production setting; that its guardrails behave as designed (neither guardrail ever raised a score, and their effect on the composite is measured); and that on 142 mature investments the full pipeline's score is rank-correlated with total-value multiples, with a modest but measurable margin over an unconstrained LLM. We do not claim that the pipeline would screen incoming deals better than analysts do, that the improvement survives without hindsight in the retrieved evidence, that the lower-only rule matters beyond what any second review would add, or that scores are stable across repeated runs. Section~\ref{sec:limits} sets out each of these limits.

\section{Related Work}

\paragraph{Machine learning for VC screening.} Prior work predicts startup success from tabular and relational data, typically using public databases such as Crunchbase \cite{arroyo2019assessment,zbikowski2021crunchbase,lyu2021gnn,cao2023gems}, and benchmarks algorithms against investor judgement \cite{retterath2020human}. Related work models business-angel decisions \cite{katcharovski2025angel} and automates pitch-deck analysis \cite{chauhan2024pitchdeck}. Survey evidence describes how VCs actually decide, with the founding team weighted most heavily \cite{gompers2020vc}, which motivates the high founder weight in our rubric. Two limitations recur: public datasets rarely contain realised fund-level returns or write-offs, and screening is usually framed as single-pass classification rather than as a staged, human-supervised workflow. Our dataset addresses the first limitation for one fund; it does not remove the selection problem discussed in \S\ref{sec:limits}.

\paragraph{LLM failure modes relevant to scoring.} LLMs exhibit sycophancy, aligning outputs with the user's or document's stance \cite{sharma2024sycophancy,perez2023discovering,natan2026sycophant}, and LLM judges show position and verbosity biases \cite{zheng2023judging}. In investment analysis specifically, LLMs show systematic preferences that can override the evidence presented \cite{lee2025yourai}. Hosted LLMs are not reproducible even at temperature zero \cite{atil2025nondeterminism}, partly because of batch-dependent numerics on shared inference servers \cite{he2025nondeterminism}. Chain-of-thought explanations can be unfaithful to the computation that produced an answer \cite{turpin2023unfaithful}. These findings motivate keeping the model out of the arithmetic and storing evidence rather than relying on regenerated explanations, in line with arguments for inherently interpretable models in high-stakes settings \cite{rudin2019stop}.

\paragraph{Look-ahead and survivorship bias in financial LLM evaluation.} Pretrained models can leak post-decision information through their weights \cite{sarkar2024lookahead,glasserman2023lookahead,lopezlira2025memorization,gao2025lookahead}; masking entity names reduces but does not reliably remove this \cite{lopezlira2025memorization}. Retrieval adds a second channel when live search engines or mutable documents are queried \cite{kong2026bias}. Kong et al.\ \cite{kong2026bias} propose a structural validity checklist (temporal sanitation, dynamic universe construction, rationale robustness, epistemic calibration and cost reporting); Appendix~\ref{app:checklist} maps our evaluation onto it. Survivorship bias, a classic problem in performance studies \cite{brown1992survivorship}, appears in our setting as \emph{selection}: only funded deals have observed outcomes.

\paragraph{Tool use and retrieval.} Retrieval-augmented generation \cite{lewis2020rag} grounds outputs in retrieved passages, and agentic patterns interleave reasoning with tool calls chosen at run time \cite{yao2023react,schick2023toolformer}. Run-time tool choice makes the evidence set itself a random variable. Our pre-fetch pattern trades flexibility for templated queries whose results are assembled before generation and logged with each screen. Processing untrusted documents also exposes LLM pipelines to indirect prompt injection \cite{greshake2023injection}.

\paragraph{Regulatory context.} The FCA describes itself as a technology-agnostic, principles-based and outcomes-focused regulator and does not impose AI-specific rules; it points instead to existing frameworks including the Principles for Businesses, the Senior Managers and Certification Regime (SM\&CR), the Senior Management Arrangements, Systems and Controls sourcebook (SYSC) and, where relevant, the Consumer Duty \cite{fca2024aiupdate}. Record-keeping obligations sit in SYSC~9 \cite{fcasysc9}. Compliance is therefore a property of a firm's processes, not of a software architecture. We describe DealsFlow's logging and approval controls as features that can \emph{support} a firm in evidencing its own systems, controls and accountability, not as making the firm compliant.

\section{System Design}
\label{sec:design}

\begin{figure*}[t]
  \centering
  \includegraphics[width=\textwidth]{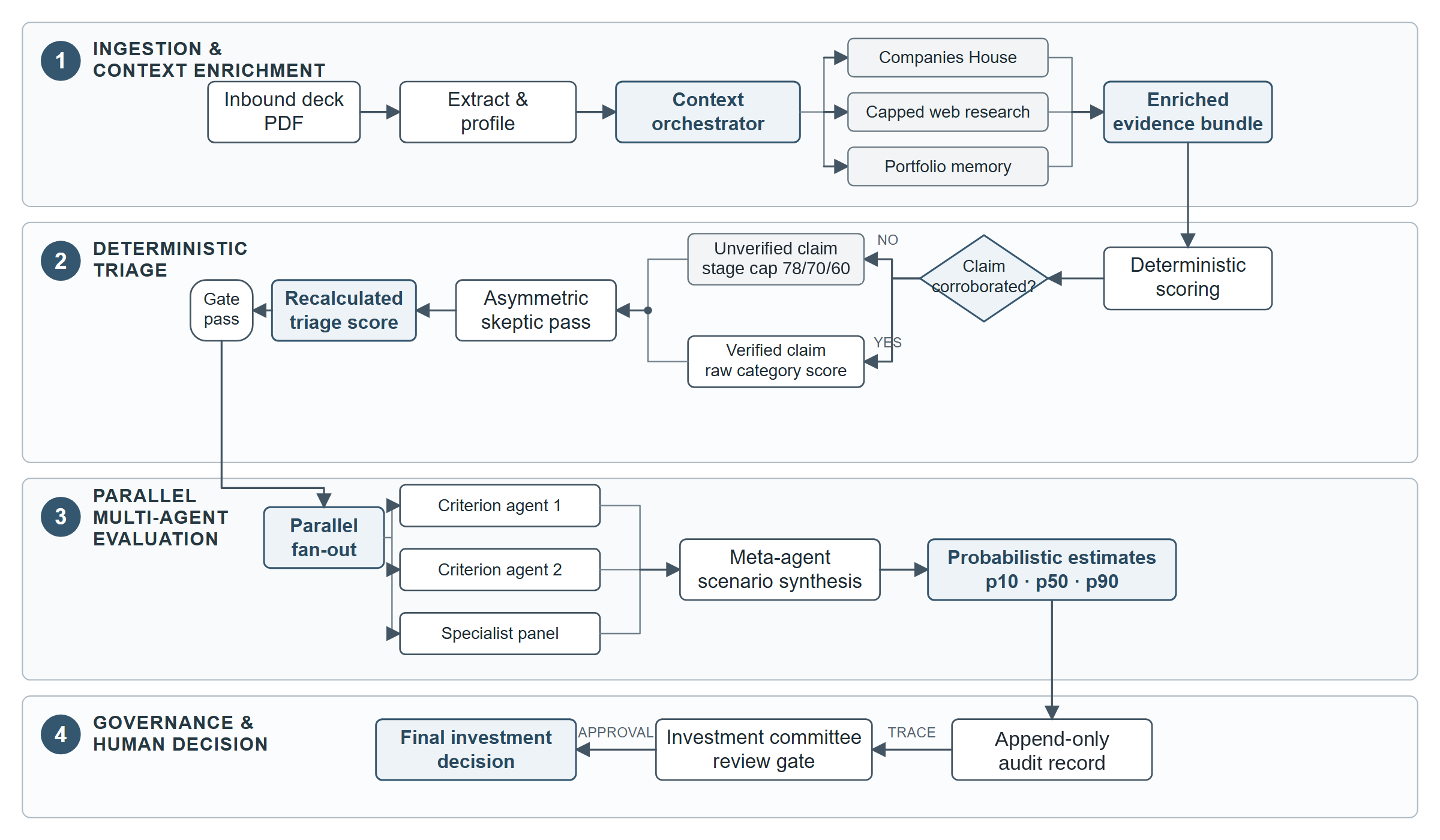}
  \caption{DealsFlow pipeline. (1) Ingestion and evidence pre-fetch; (2) deterministic triage with the corroboration ceiling and the lower-only sceptic pass; (3) parallel per-criterion evaluation for applications that pass the gate; (4) audit record and mandatory human approval. The ceiling values are stage-dependent (\S\ref{sec:design}). Components in stage 3 and the portfolio-memory branch of stage 1 are described in \S\ref{sec:other} but not evaluated in this paper.}
  \Description{Flow diagram in four horizontal bands. Band 1: an inbound PDF deck is extracted and profiled, then a context orchestrator queries Companies House, capped web research and portfolio memory and produces an enriched evidence bundle. Band 2: deterministic scoring, then a decision on whether each claim is corroborated. Uncorroborated claims are capped at a stage-dependent ceiling of 78, 70 or 60; corroborated claims keep the raw category score. Both paths feed an asymmetric sceptic pass that can only lower scores, then a recalculated triage score and a gate. Band 3: a parallel fan-out to criterion agents and a specialist panel, a meta-agent that synthesises scenarios, and probabilistic estimates at p10, p50 and p90. Band 4: an append-only audit record, an investment committee review gate, and a final investment decision reached only through human approval.}
  \label{fig:arch}
\end{figure*}

Figure~\ref{fig:arch} shows the pipeline. The system runs in production at Sapphire Capital Partners, a UK early-stage fund manager; this section describes the current build, which is the one evaluated here (Appendix~\ref{app:config} lists models and settings).

\subsection{Extraction and corroboration}

An analyst uploads up to 10 PDF decks per batch. A worker extracts text (pypdf) and runs one LLM pass that produces a structured profile (company name, sector, stage, capital ask, revenue band, geography, founders and their roles). The harness then gathers external evidence: a Companies House profile snapshot (status, incorporation date, charges) and active-officer list \cite{companieshouseapi}, and a templated set of at most 10 web queries. There is one query per founder (up to three), plus a second where a role is stated; three company queries (news and funding, competitors, customers and traction); and one market-size query when a sector was extracted. Queries go through a metasearch library (\texttt{ddgs}) that rotates between several search backends, returning five results per query; the top two pages per query are crawled and truncated to 4{,}000 characters. Every retrieved page is assigned a numbered source index.

Corroboration is recorded per category, not per claim. When scoring, the model labels the support for each category as \emph{external} (a numbered web source backs it), \emph{registry} (Companies House data backs it), \emph{deck\_only} (only the deck asserts it) or \emph{none}. Code fails closed: only \emph{external} and \emph{registry} count as corroborated. The label is itself a model judgement, not a code-level match between a claim and a source; \S\ref{sec:pit} tests whether it is contaminated by hindsight.

\subsection{Deterministic composite and the corroboration ceiling}

The LLM assigns each of six categories a score $s_i \in [0,100]$ against one set of score bands shared by all categories and a one-line definition of each (Appendix~\ref{app:rubrics}). It is told to judge against what is normal for the company's stage, to use the full range, and not to deduct for missing corroboration, because the ceiling below is applied separately in code. The weights are vision 1.0, market 1.5, financials and valuation 1.5, founders 2.0, traction 1.5 and differentiation 1.5. The weights were agreed with the fund's investment committee, with founders weighted highest, consistent with survey evidence that venture capitalists weight the team most heavily \cite{gompers2020vc}. They have not changed since they were first committed and were not tuned on the evaluation portfolio. The composite is computed in code:
\begin{equation}
  S = \frac{\sum_{i=1}^{6} w_i\, \tilde{s}_i}{\sum_{i=1}^{6} w_i}, \qquad \textstyle\sum_i w_i = 9.0 .
\end{equation}
For the four factual categories (market, financials, traction and differentiation), the effective score is
\begin{equation}
  \tilde{s}_i = \begin{cases} s_i & \text{if category } i \text{ is labelled \emph{external} or \emph{registry},} \\ \min(s_i, c_\tau) & \text{otherwise,} \end{cases}
\end{equation}
where $c_\tau$ depends on the deck's stage tier $\tau$: $c_{\text{early}}=78$, $c_{\text{unknown}}=70$ and $c_{\text{growth}}=60$. Code assigns $\tau$ by keyword match on the stage the deck reports: growth terms such as ``Series A'' are checked first, then early-stage terms such as ``seed'' or ``pre-revenue''; a missing or unrecognised stage is ``unknown''. Founders and vision are exempt because they are judgement-based and rarely externally checkable. The ceiling scales with stage because public evidence is a reasonable expectation of a growth-stage company but not of a pre-seed one, and a flat ceiling would make the composite a proxy for public-data availability. It is a ceiling, not a multiplier: it only affects a category that claims a high score without evidence, and scores already below it are untouched. The values were set before this study and were not tuned on the evaluation portfolio.

\begin{proposition}
\label{prop:ceiling}
If no factual category is corroborated, then $S \le (1.0\cdot 100 + 2.0\cdot 100 + 6.0\cdot c_\tau)/9$, which is 85.3, 80.0 and 73.3 for $c_\tau=78$, 70 and 60, regardless of the vision and founder scores.
\end{proposition}

The ceiling is therefore a mechanical bound, not a finding. It limits how high a deal with little external evidence can score, so \S\ref{sec:evaldesign} tests whether evidence availability differs by outcome, and compares the full system with a baseline that uses corroboration counts alone.

\subsection{Asymmetric sceptic pass}

After initial scoring, a second call to the same model through the same evaluation path (the \emph{sceptic}) receives each category's score and cited evidence and returns a proposed revision. Code accepts a revision only if it is lower, and limits each reduction to 20 points:
\begin{equation}
  \tilde{s}_i^{\text{new}} = \max\!\left(\tilde{s}_i - 20,\; \min(\tilde{s}_i, s_i^{\text{sceptic}})\right).
\end{equation}
The composite is then recomputed. The sceptic is instructed to adjust a score only where it can name a specific contradiction or implausibility, not merely because corroboration is missing, since the ceiling already handles that and a second deduction would double-penalise.

\subsection{Declared-dependency pre-fetch}

In a run-time tool-calling agent \cite{yao2023react,schick2023toolformer}, the model decides during generation which searches to run, so two runs on the same deck can see different evidence. DealsFlow instead requires each evaluation step to declare its data dependencies (registry lookup, a fixed set of templated queries, deck passages) before generation. The harness retrieves them, assigns each source a numbered index, and then makes a single generation call over that context. The source list (title and URL) is stored with the screen.

This does not make the LLM deterministic. All calls use temperature 0.1 with no fixed seed, and server-side numerics vary even at temperature zero \cite{atil2025nondeterminism,he2025nondeterminism}. What it does guarantee is narrower and, for audit purposes, more useful: (a) the queries are fixed by template rather than chosen by the model, and the evidence is assembled before generation, so the model cannot change what it retrieves while answering; and (b) because the source list, the category scores before and after each guardrail and the model identifier are stored, the decision record can be inspected later without regenerating anything. Triage does not store full page text or the full prompt, so a triage screen can be audited at the level of sources, not passages; stage evaluations additionally store the full prompt, the response and a prompt hash. Pre-fetch does not make the evidence identical across screens of the same deck: search backends rotate, pages change, and search results are cached for only 24 hours. We did not measure run-to-run variance (\S\ref{sec:limits}).

\subsection{Sequential gating}

Funds configure pipeline stages in an admin console. Each stage has a type (rule-based check, LLM evaluation, specialist panel, or human decision gate), criteria, weights and a pass mark. Cheap stages run first, so the expensive parallel evaluation in stage 3 runs only on applications that pass triage. Intake fields carry \emph{semantic roles} (e.g.\ revenue band, sector, geography) so that analytics work across funds with different form designs.

\subsection{Components not evaluated in this paper}
\label{sec:other}

The following components are deployed but are outside the scope of the experiments; we describe them for completeness and make no claims about their effectiveness.

\paragraph{Parallel criterion agents (stage 3).} Using LangGraph's \texttt{Send()} primitive, one sub-agent per stage criterion runs in parallel with its own pre-fetched evidence; a synthesis step produces p10/p50/p90 survival estimates.

\paragraph{Factor-based similarity.} New applications are compared with past deals over seven semantic roles with weights sector 1.0, stage 0.8, business model 0.6, revenue band 0.6, geography 0.5, summary 0.5 and raise amount 0.4. Let $R_{ij}$ be the roles populated for both deals $i$ and $j$, and $s_r \in [0,1]$ the per-role similarity. Then
\begin{align}
  S_{ij} &= \frac{\sum_{r\in R_{ij}} w_r\, s_r}{\sum_{r\in R_{ij}} w_r}, \\
  \mathrm{sim}(i,j) &= \mathbf{1}\!\left[|R_{ij}| \ge 2\right]\mathbf{1}\!\left[S_{ij} \ge 0.60\right] S_{ij}^{\gamma}\, 0.5^{\delta/h},
\end{align}
with $\gamma=2$ and half-life $h=730$ days; the age $\delta$ runs from the past deal's decision date (or its deployment date if there is none) to the as-of date. Per-role similarity is an exact match for sector and geography; ordinal, $1-|\ell_i-\ell_j|/(L-1)$ over $L$ levels, for stage and revenue band; $1-|x_i-x_j|/\max(|x_i|,|x_j|,1)$ for raise amount; and token Jaccard similarity for business model and business summary. No embeddings are used.

\paragraph{Bounded historical adjustment.} Matched past deals with verified outcomes are coded acquired/IPO $=1.0$, active $=0.4$ scaled by $\min(1, \text{age}/1095\text{ days})$ so that a young holding counts for less, pivoted $=0.0$ and failed $=-1.0$; uninvested and rejected deals are excluded because their outcomes are unobserved. Let $b\in[-1,1]$ be the similarity-weighted mean code and $\lambda=\min(1, \sum_j \mathrm{sim}_j/3)$ a support weight, computed only when at least three funded matches exist. Then
\begin{equation}
  p_{\text{adj}} = \min\!\left(1, \max\!\left(0,\; p_{\text{model}} + 0.35\,\lambda\, b\right)\right),
\end{equation}
so the shift is at most $\pm 0.35$. Note that $b$ is an outcome index, not a probability, so this is a bounded heuristic nudge rather than a probabilistic update. It never changes category scores. When $p_{\text{model}}$ recommends approval but $b \le -0.5$ with $\lambda \ge 0.5$, the system raises a divergence flag for the committee instead.

\paragraph{Monitoring and reflection.} After investment, quarterly founder updates produce new survival snapshots. When an outcome is logged, a reflection agent proposes lessons, which stay inactive until supported by at least three distinct outcomes and are then shown to later evaluations as cited hypotheses.

\section{Governance Design}
\label{sec:governance}

\paragraph{Audit record.} Each triage screen stores, on the screened-deck record, the structured profile, the category scores before and after each guardrail, the numbered source list (title and URL) and the model identifier. Stage evaluations in the full pipeline additionally store the full prompt and response and a prompt hash. Triage does not store full page content or the full prompt. Records are append-only by application design and isolated per tenant by row-level security; no cryptographic tamper-evidence (such as hash chaining) is used.

\paragraph{Human authority.} The system may automatically reject an application below a configured threshold, but cannot approve: progression to investment requires an explicit human action at a terminal investment-committee stage that cannot be removed from a pipeline. Overriding a system recommendation requires a written justification, stored with the user identity and timestamp.

\paragraph{How this relates to regulatory obligations.} These controls are designed to help a firm evidence the things the FCA's existing framework asks of it: clear senior-management accountability for decisions (SM\&CR), adequate systems and controls (SYSC), and adequate records (SYSC~9) \cite{fca2024aiupdate,fcasysc9}. They do not by themselves make a firm compliant, and we make no such claim.

\paragraph{Untrusted input.} Founder decks and web pages are untrusted and can carry indirect prompt injections \cite{greshake2023injection}. Defences consist of delimiter wrapping of extracted text, pattern-based stripping of instruction-like strings, and model-reported injection flags. They were not evaluated here. All similarity search, lesson retrieval and base rates are computed within one tenant's data.

\section{Evaluation Design}
\label{sec:evaldesign}

\subsection{Portfolio dataset}

The dataset is a portfolio record provided by Sapphire Capital Partners, a UK early-stage fund manager: 245 company records exported from the manager's portfolio-administration system on 8 July 2026, which we use as the valuation date. Two records have no recorded investment amount (one with zero value, one with a residual value of \pounds382) and are excluded as data-entry artefacts, leaving $N=243$ SEIS/EIS investments made between September 2015 and April 2026; none is a VCT holding. These represent \pounds272.50m invested and \pounds417.22m of total value (realised proceeds of \pounds46.4m plus \pounds370.8m of unrealised value at the latest marks), an aggregate multiple of 1.53$\times$. Stages at entry are seed (168), pre-seed (73) and Series~A (2).

Each company is assigned to one of eight outcome classes from its total-value multiple and exit flag (Table~\ref{tab:portfolio}): written off if total value is zero; otherwise fully exited at a profit, at cost, or at a loss if all shares have been exited; otherwise active breakout ($\ge 3\times$), up-valued, held at cost (within $\pm$0.5\% of cost) or impaired. Holding period runs from the first investment date to the exit date, or to the valuation date for active holdings. Value is concentrated: among the 196 positions with non-zero value, the top 10 account for 50.6\% of total value and the top 49 for 86.5\% (Table~\ref{tab:concentration}); the Gini coefficient over these 196 positions is 0.77 (0.81 over all 243).

\begin{table*}[t]
  \caption{Portfolio by outcome class ($N=243$, valuation date 8 July 2026). Holding period is the mean, in years, from first investment to exit, or to the valuation date for active holdings. Aggregate multiple is total value over capital invested within the class.}
  \label{tab:portfolio}
  \begin{tabular}{lrrrr}
    \toprule
    Outcome class & $n$ & Holding period (yrs) & Aggregate multiple & Individual range \\
    \midrule
    Active breakout           & 15  & 5.50 & 4.61$\times$ & 3.12--14.74$\times$ \\
    Active up-valued          & 50  & 4.06 & 1.67$\times$ & 1.09--2.80$\times$ \\
    Active held at cost       & 102 & 1.94 & 1.00$\times$ & 1.00$\times$ \\
    Active impaired           & 14  & 4.41 & 0.54$\times$ & 0.04--0.99$\times$ \\
    Fully exited (profitable) & 3   & 2.15 & 4.44$\times$ & 1.04--5.81$\times$ \\
    Fully exited (at cost)    & 2   & 1.80 & 1.00$\times$ & 1.00$\times$ \\
    Fully exited (loss)       & 10  & 4.64 & 0.30$\times$ & 0.00--0.49$\times$ \\
    Written off               & 47  & 3.42 & 0.00$\times$ & 0.00$\times$ \\
    \addlinespace[4pt]
    Total                     & 243 & 3.14 & 1.53$\times$ & 0.00--14.74$\times$ \\
    \bottomrule
  \end{tabular}
\end{table*}

\begin{table}[t]
  \caption{Value concentration among the 196 positions with non-zero total value. Shares are of total portfolio value (\pounds417.22m, realised plus unrealised).}
  \label{tab:concentration}
  \begin{tabular}{lrrr}
    \toprule
    Bracket & Deals & Value (\pounds m) & Share of total value \\
    \midrule
    Top 1\%  & 2  & 74.1  & 17.8\% \\
    Top 5\%  & 10 & 211.1 & 50.6\% \\
    Top 10\% & 20 & 274.7 & 65.8\% \\
    Top 25\% & 49 & 360.8 & 86.5\% \\
    \bottomrule
  \end{tabular}
\end{table}

\begin{figure}[t]
  \centering
  \includegraphics[width=\columnwidth]{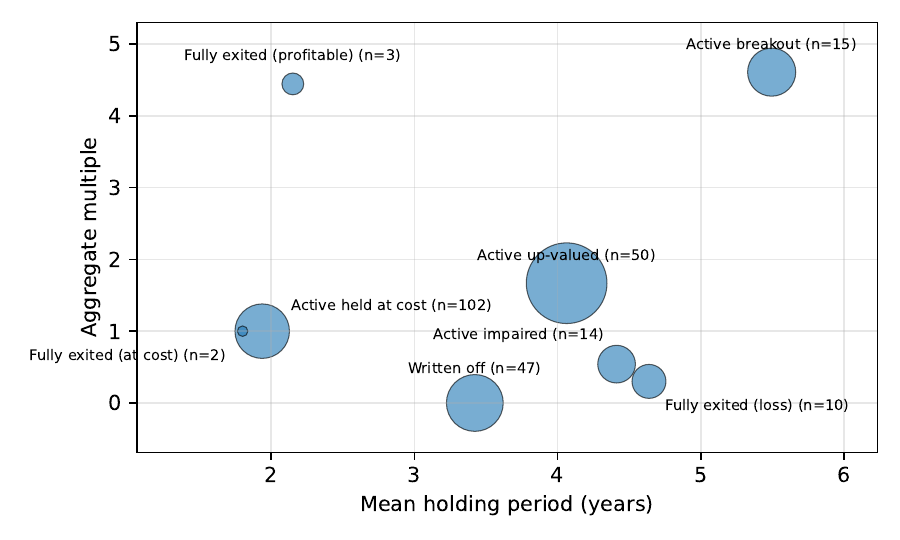}
  \caption{Mean holding period against aggregate multiple by outcome class; bubble area is proportional to capital invested. Generated from the same records as Table~\ref{tab:portfolio}.}
  \Description{Bubble chart. The horizontal axis is mean holding period in years, from about 1.8 to 5.5, and the vertical axis is aggregate multiple, from 0 to about 4.6 times. Active breakout is top right at about 5.5 years and 4.6 times. Fully exited profitable is top left at about 2.2 years and 4.4 times. Active up-valued is near 4 years and 1.7 times and is the largest bubble. Active held at cost and fully exited at cost sit at 1 times, under 2 years. Active impaired and fully exited loss sit below 1 times at about 4.4 and 4.6 years. Written off sits at 0 times at about 3.4 years.}
  \label{fig:holding}
\end{figure}

\paragraph{Outcome maturity.}
\label{sec:maturity}
Outcome labels depend on how long a company has had to succeed or fail. Table~\ref{tab:maturity} shows that none of the 73 investments made within two years of the valuation date has been written off, and 81\% of them are still held at cost, typically because they have not yet been revalued. A score that merely tracks company age would therefore appear to identify write-offs: a logistic regression on entry year alone reaches AUC$_0=0.72$ [0.65, 0.78] on all 243 deals. We use the 142 investments made at least three years before the valuation date (43 write-offs, 22 above $2\times$) as the primary evaluation set. On this set entry year showed no clear discrimination (AUC$_0=0.45$ [0.34, 0.55]). All 142 entered at seed stage.

\begin{table}[t]
  \caption{Outcomes by time since first investment (valuation date 8 July 2026).}
  \label{tab:maturity}
  \begin{tabular}{lrrrr}
    \toprule
    Years since investment & $n$ & Written off & Held at cost & $>2\times$ \\
    \midrule
    $<2$  & 73 & 0 (0\%)   & 59 (81\%) & 4 \\
    2--3  & 28 & 4 (14\%)  & 17 (61\%) & 0 \\
    3--5  & 93 & 28 (30\%) & 24 (26\%) & 7 \\
    $\ge 5$ & 49 & 15 (31\%) & 2 (4\%)  & 15 \\
    \bottomrule
  \end{tabular}
\end{table}

\paragraph{What this dataset can and cannot test.} Every deal in the dataset was approved by the fund's analysts and committee. The evaluation therefore measures whether DealsFlow \emph{ranks funded deals by eventual outcome}, a harder, conditional task than separating good from bad applications in raw deal flow. It cannot measure false negatives (good companies the system would have rejected), because rejected applications have no observed outcome. We return to this in \S\ref{sec:limits}.

\subsection{Research questions}

\begin{description}
  \item[RQ1 (leakage)] Does external corroboration availability differ by outcome, and how much of the score's discrimination does corroboration alone explain?
  \item[RQ2 (discrimination)] On the 142 mature deals, does the composite rank outcomes, and how does it compare with simple baselines?
  \item[RQ3 (ablation)] What does each guardrail contribute?
  \item[RQ4 (analyst time)] What is the total analyst time per application with DealsFlow, from opening the generated report to a completed assessment, compared with manual triage?
  \item[RQ5 (factual support)] Do the cited sources support the claims in the generated reports, are important issues missed, and how much correction do analysts still have to make?
\end{description}

\subsection{Leakage channels}
\label{sec:pit}

Let $t_j$ be the investment-committee date for deal $j$. A backtest should give the system only information available at $t_j$. Our retrieval does not enforce this, so for each channel we state what is and is not controlled, and \S\ref{sec:leak} measures the consequences. We do not claim point-in-time isolation.

\paragraph{Deck.} The application file as submitted before $t_j$, from the fund's archive. No later investor updates were included.

\paragraph{Companies House.} The API returns current data. The harness passes the current company status, incorporation date, charge counts and active officers to the model without a date filter, so this channel can carry post-$t_j$ information (for example a dissolved status). We report this as residual leakage (\S\ref{sec:limits}).

\paragraph{Web search.} Live search engines cannot be restricted to a historical index, and dated pages can be edited later \cite{kong2026bias}. Web queries carried no date restriction in any run, so retrieved evidence may include information published after $t_j$; we treat this as residual leakage and report it as such.

\paragraph{Model weights.} The extraction model may already know what happened to some companies \cite{sarkar2024lookahead,lopezlira2025memorization}. The main evaluation uses GPT-6 Luna, accessed through Microsoft Azure, whose stated knowledge cutoff is 18 May 2026 \cite{openai2026gpt6luna}. This cutoff falls after most outcomes in our portfolio were determined, so the model may know what happened to some companies.  The harness has no control for this channel, and we report it as residual leakage (\S\ref{sec:limits}).

\subsection{Configurations and baselines}

\begin{description}
  \item[Unconstrained] One LLM call reads the deck and evidence and outputs the six category scores; code computes the composite; no ceiling, no sceptic.
  \item[Ceiling only] As above, with the corroboration ceiling.
  \item[Full] Ceiling plus sceptic.
  \item[Sceptic only] Sceptic without ceiling, to separate the two guardrails.
  \item[Corroboration-only] No score judgement: $S$ is the fraction of the four factual categories labelled \emph{external} or \emph{registry}. If this matches the full pipeline, the LLM's judgement adds little beyond evidence availability.
  \item[Vintage only] Logistic regression on entry year, to measure what time-at-risk alone explains.
  \item[Structured-field LR] Logistic regression predicting write-off from sector, stage and initial cheque size, with 5-fold cross-validated out-of-fold predictions. The initial cheque is the fund's first-round subscription. We do not use total capital invested, because follow-on rounds are decided after $t_j$ and go disproportionately to companies that are doing well (breakouts received a mean of 3.2 funding rounds from the fund, against 1.4 for holdings at cost): only 48.5\% of capital was invested in first rounds.
  \item[Initial cheque] The first-round cheque size alone, as a proxy for the fund's own conviction at the time of investment.
\end{description}

All LLM configurations use the same model, evidence context and rubric (Appendix~\ref{app:config}).

\subsection{Metrics and statistics}

Our primary metric is Spearman's $\rho$ between composite score and total-value multiple, that is, realised proceeds plus current marks over capital invested; many marks on active holdings are the manager's valuations rather than realised returns. Secondary metrics are two AUCs and per-class means with standard deviations. AUC$_0$ is the probability that a randomly chosen deal that was not written off scores higher than a randomly chosen write-off; AUC$_2$ is the probability that a randomly chosen deal with multiple $>2\times$ scores higher than a randomly chosen other deal. Intervals are 95\% percentile bootstrap intervals over deals with 10{,}000 resamples \cite{efron1994bootstrap}; differences between configurations are tested with a paired bootstrap on the same deals. All investment-outcome results use the 142 mature deals; the analyst measures use the samples described in \S\ref{sec:analyst}. With 142 deals and 43 write-offs, the half-width of a 95\% interval on AUC$_0$ is roughly $\pm 0.10$ (Hanley--McNeil approximation at AUC $\approx 0.65$), so differences smaller than that are inconclusive; paired comparisons on the same deals are more sensitive.

\subsection{Analyst time and factual-support protocol}
\label{sec:analyst}

Investment outcomes cannot show whether the harness helps an analyst, so we add two measures. Both use 20 inbound applications that are separate from the historical portfolio and have no known outcomes. Two analysts took part. Each application was triaged once manually and once with DealsFlow, and for each application one analyst performed the manual triage while the other used DealsFlow, so neither analyst processed the same application twice. Assignments were counterbalanced across the 20 applications, with each analyst doing 10 applications in each workflow.

\paragraph{Total analyst time.} We measure the time from opening a generated report to a completed final assessment. This includes reading the report, verifying its sources, correcting errors and completing the assessment, not only the machine processing time. The comparison is with manual triage of the same application, timed from opening the deck to a completed summary. Machine time, which runs without an analyst, is recorded separately. The four assisted components (reading the report, verifying sources, correcting errors and completing the final assessment) were logged separately.

\paragraph{Factual-support quality.} Two reviewers independently rated every distinct claim that cites a numbered source in each of the 20 generated reports: 240 claims, so 480 rating decisions before reconciliation. For each claim the reviewer opens the source and records whether it supports the claim: supported, partly supported, not supported, or source not accessible. For the four ceiling-eligible categories of each report (80 labels), reviewers also record whether an \emph{external} or \emph{registry} label is confirmed by a cited source, and whether a category labelled \emph{deck\_only} or \emph{none} could have been corroborated by a source the harness missed. They list material issues the report does not mention, such as contradictions with filings, and count the corrections made before the assessment was completed. The two reviewers are the two analysts who took part in the timing. They rated every claim independently before any discussion, then discussed disagreements against the cited sources, and a third reviewer adjudicated unresolved cases. The third reviewer is one of the authors, who manages the partnership between the university and the fund manager. We report agreement between the two reviewers on their initial ratings, calculated before reconciliation, as the share of claims rated alike and as Cohen's $\kappa$; the support counts we report are after reconciliation.

\paragraph{Statistics.} With 20 applications we report means with standard deviations and 95\% confidence intervals for the mean, and a paired difference for the timing comparison. Intervals are percentile bootstrap intervals from 10{,}000 resamples of applications, and the resampling keeps each application's manual and assisted times together when estimating the saving. We make no significance claims.

\section{Results}
\label{sec:results}

\subsection{RQ1: Is the signal evidence availability?}
\label{sec:leak}

\begin{table}[t]
  \caption{Corroboration by outcome class (142 mature deals). Rate is the pooled share of the four factual categories per deal (market, financials, traction, differentiation) labelled \emph{external} or \emph{registry}; $n$ is the number of deals, so each class has $4n$ categories. A Kruskal--Wallis test on the 142 per-deal corroboration fractions, rather than on the 568 categories as if independent, gives $H(4)=10.65$, $p=0.031$.}
  \label{tab:leak}
  \begin{tabular}{lrrr}
    \toprule
    Outcome class & $n$ & Corroborated / total & Rate \\
    \midrule
    $>2\times$                      & 22  & 56 / 88   & 63.6\% \\
    $>1\times$ to $2\times$         & 31  & 54 / 124  & 43.5\% \\
    At cost                         & 27  & 50 / 108  & 46.3\% \\
    Impaired / loss (excl.\ write-offs) & 19  & 35 / 76   & 46.1\% \\
    Written off                     & 43  & 65 / 172  & 37.8\% \\
    \addlinespace[4pt]
    Overall                         & 142 & 260 / 568 & 45.8\% \\
    \bottomrule
  \end{tabular}
\end{table}

Corroboration availability differs by outcome (Kruskal--Wallis $p=0.031$). Investments that went on to exceed $2\times$ had 63.6\% of their factual categories corroborated, against 37.8\% for write-offs; the three middle classes lie between, at 43--46\%. This is the pattern a hindsight-contaminated retrieval would produce, and it is also the pattern that genuine early traction would produce. With undated search we cannot tell which, and we do not claim to.

Two results limit how much corroboration alone can explain. First, a baseline that scores each deal only by its fraction of corroborated categories reaches $\rho=0.22$ [0.05, 0.37] and AUC$_0=0.61$ [0.50, 0.70] (Table~\ref{tab:main}), well below the full pipeline; the paired difference in $\rho$ is $+0.235$ [0.037, 0.429] (Table~\ref{tab:paired}). Second, the stage-aware ceiling, the mechanism that depends on the corroboration labels, is weak: it changes the composite of 48.6\% of deals, by 1.23 points on average (Table~\ref{tab:guard}), and the ceiling-only configuration is almost identical to the unconstrained LLM ($\rho=0.384$ against $0.381$). The ceiling therefore accounts for little of the overall discrimination, which comes mostly from the model's category scores. But those scores are produced from the same undated evidence, so the corroboration gap remains a plausible route for hindsight, and the results below should be read with that caveat.

\subsection{RQ2--RQ3: Discrimination and ablation (142 mature deals)}
\label{sec:main}

\begin{table*}[t]
  \caption{Results on the 142 mature deals. 95\% bootstrap intervals (10{,}000 resamples) in brackets. AUC$_{0}$: write-off vs.\ rest; AUC$_{2}$: multiple $>2\times$ vs.\ rest. The corroboration-only row uses the corroboration labels from the scoring runs; the vintage, structured-field and initial-cheque rows are computed from the portfolio records alone. Each deal was scored once per configuration.}
  \label{tab:main}
  \begin{tabular}{lrrrr}
    \toprule
    Configuration & Spearman $\rho$ & AUC$_{0}$ & AUC$_{2}$ & Mean score, write-offs / $>2\times$ \\
    \midrule
    Unconstrained LLM        & 0.381 [0.239, 0.512] & 0.681 [0.588, 0.768] & 0.727 [0.608, 0.838] & 67.5 / 79.8 \\
    Ceiling only             & 0.384 [0.238, 0.517] & 0.681 [0.586, 0.768] & 0.751 [0.628, 0.864] & 66.9 / 78.4 \\
    Sceptic only             & 0.435 [0.301, 0.555] & 0.705 [0.616, 0.787] & 0.762 [0.651, 0.864] & 63.3 / 78.6 \\
    Full (ceiling + sceptic) & 0.451 [0.317, 0.570] & 0.709 [0.620, 0.790] & 0.791 [0.678, 0.892] & 62.7 / 77.2 \\
    \addlinespace[4pt]
    Corroboration-only       & 0.216 [0.048, 0.372] & 0.605 [0.503, 0.702] & 0.696 [0.579, 0.804] & -- \\
    Vintage only             & $-$0.01 [$-$0.19, 0.17] & 0.45 [0.34, 0.55] & 0.64 [0.50, 0.77] & -- \\
    Structured-field LR      & 0.27 [0.11, 0.43] & 0.64 [0.53, 0.75] & 0.68 [0.56, 0.79] & -- \\
    Initial cheque           & 0.03 [$-$0.13, 0.20] & 0.51 [0.41, 0.62] & 0.65 [0.52, 0.76] & -- \\
    Random                   & 0.00 & 0.50 & 0.50 & -- \\
    \bottomrule
  \end{tabular}
\end{table*}

The baselines set the bar. Sector, stage and initial cheque size rank mature outcomes with $\rho=0.27$. Every mature deal entered at seed stage and cheque size alone showed little discrimination, but that does not show where the combined model's signal comes from: attributing it to sector would need a feature ablation, which we did not run. The fund's cheque size, a proxy for its conviction at the time, showed no clear separation of write-offs (AUC$_0=0.51$). 

\paragraph{DealsFlow against the baselines.} The unconstrained LLM already reaches $\rho=0.381$ [0.239, 0.512] and AUC$_0=0.681$; the full pipeline reaches $\rho=0.451$ [0.317, 0.570], AUC$_0=0.709$ [0.620, 0.790] and AUC$_2=0.791$ [0.678, 0.892]. The point estimates of all four LLM configurations are above those of the baselines built from the portfolio records alone, including the structured-field regression ($\rho=0.27$, AUC$_0=0.64$, AUC$_2=0.68$). The intervals overlap, however, and we did not run a paired test against the regression, so we do not claim a difference.

\begin{table*}[t]
  \caption{Paired bootstrap differences between configurations on the same 142 deals (10{,}000 resamples; 95\% intervals in brackets).}
  \label{tab:paired}
  \begin{tabular}{lrrr}
    \toprule
    Comparison & Difference in $\rho$ & Difference in AUC$_0$ & Difference in AUC$_2$ \\
    \midrule
    Full $-$ Unconstrained      & $+$0.070 [0.035, 0.105] & $+$0.027 [0.006, 0.050] & $+$0.064 [0.041, 0.091] \\
    Full $-$ Ceiling only       & $+$0.067 [0.034, 0.103] & $+$0.028 [0.006, 0.051] & $+$0.040 [0.015, 0.067] \\
    Full $-$ Sceptic only       & $+$0.016 [$-$0.006, 0.040] & $+$0.004 [$-$0.008, 0.016] & $+$0.029 [0.011, 0.051] \\
    Full $-$ Corroboration-only & $+$0.235 [0.037, 0.429] & $+$0.104 [$-$0.034, 0.242] & $+$0.095 [$-$0.037, 0.230] \\
    \bottomrule
  \end{tabular}
\end{table*}

\paragraph{Ablation.} The full pipeline improves on the unconstrained LLM on all three metrics, with paired intervals that exclude zero (Table~\ref{tab:paired}), but the gains are small. Most of the gain is already present in the sceptic-only configuration. That configuration also uses a second call, so this does not show that the lower-only rule is responsible (\S\ref{sec:limits}). Adding the ceiling to it changes $\rho$ by $+0.016$ [$-0.006$, 0.040] and AUC$_0$ by $+0.004$ [$-0.008$, 0.016], neither distinguishable from zero, and improves AUC$_2$ by $+0.029$ [0.011, 0.051]. Adding the sceptic to the ceiling-only configuration is worth $+0.067$ in $\rho$ [0.034, 0.103]. Against the corroboration-only baseline, the full pipeline is higher in $\rho$ ($+0.235$ [0.037, 0.429]); the differences in AUC$_0$ and AUC$_2$ are not distinguishable from zero.

\begin{table}[t]
  \caption{Composite score of the full pipeline by outcome group (142 deals).}
  \label{tab:dist}
  \begin{tabular}{lrrr}
    \toprule
    Outcome group & Mean $\pm$ SD & Median & Q1--Q3 \\
    \midrule
    $>2\times$                  & 77.2 $\pm$ 8.3  & 78.6 & 73.0--81.6 \\
    $>1\times$ to $2\times$     & 71.4 $\pm$ 7.4  & 71.3 & 67.5--76.8 \\
    At cost                     & 69.2 $\pm$ 9.3  & 70.9 & 60.8--76.2 \\
    Impaired / loss             & 64.4 $\pm$ 13.3 & 65.0 & 55.2--78.1 \\
    Written off                 & 62.7 $\pm$ 11.0 & 65.2 & 58.0--70.3 \\
    \bottomrule
  \end{tabular}
\end{table}

\paragraph{Score distributions.} Group means fall steadily from 77.2 for investments above $2\times$ to 62.7 for write-offs (Table~\ref{tab:dist}), but the distributions overlap heavily. The interquartile ranges of the at-cost, impaired/loss and written-off groups overlap almost entirely, and the medians of the last two are essentially equal (65.0 and 65.2). The separation is concentrated between the best outcomes and the rest, so the scores should not be read as a fine-grained ranking.

\begin{table}[t]
  \caption{Effect of the guardrails on the 142 composite scores.}
  \label{tab:guard}
  \small
  \begin{tabular}{p{0.60\columnwidth}r}
    \toprule
    Measure & Result \\
    \midrule
    Deals affected by the ceiling & 69/142 (48.6\%) \\
    Mean reduction from the ceiling, all deals & 1.23 points \\
    Deals affected by the sceptic & 135/142 (95.1\%) \\
    Mean further reduction from the sceptic, all deals & 2.99 points \\
    Score increases from either guardrail & 0 \\
    \bottomrule
  \end{tabular}
\end{table}

\paragraph{Guardrail effects.} The ceiling changed the composite of fewer than half of the deals, by about one point on average; the sceptic changed almost all of them, by about three points. Neither increased any score, as the design requires (Table~\ref{tab:guard}). The ablation suggests the ceiling accounts for little of the overall improvement. This bears on the concern that a flat ceiling made the result arithmetic, but we did not examine the ceiling's effect on rankings within particular groups.

\subsection{RQ4: Analyst time}
\label{sec:time}

Manual triage of the 20 applications (deck reading, Companies House check, founder and competitor search, written summary) averaged 44.69 minutes per application. DealsFlow took 6.13 minutes of machine time, which runs without an analyst. The analyst time that matters is what remains afterwards: reading the generated report, verifying its sources, correcting errors and completing the final assessment. Table~\ref{tab:time} reports it.

\begin{table*}[t]
  \caption{Time per application in minutes, 20 inbound applications (mean $\pm$ standard deviation, with the 95\% confidence interval for the mean). Elapsed time adds the machine time to the analyst time, assuming the analyst starts when the machine run ends. The saving is the paired difference between manual triage and total analyst time with DealsFlow, for the same application and different analysts.}
  \label{tab:time}
  \begin{tabular}{lrr}
    \toprule
    Measure & Mean $\pm$ SD & 95\% CI for mean \\
    \midrule
    Manual triage                                  & 44.69 $\pm$ 11.02 & 40.05--49.53 \\
    DealsFlow machine processing, unattended       & 6.13 $\pm$ 1.58   & 5.51--6.85 \\
    Analyst: reading the report                    & 4.63 $\pm$ 1.26   & 4.10--5.17 \\
    Analyst: verifying sources                     & 7.04 $\pm$ 1.84   & 6.27--7.86 \\
    Analyst: correcting errors                     & 2.41 $\pm$ 1.31   & 1.85--2.96 \\
    Analyst: completing the final assessment       & 4.77 $\pm$ 1.14   & 4.30--5.27 \\
    Total analyst time with DealsFlow              & 18.85 $\pm$ 4.47  & 16.93--20.78 \\
    Elapsed time, machine plus analyst             & 24.98 $\pm$ 5.04  & 22.87--27.12 \\
    Paired analyst-time saving                     & 25.84 $\pm$ 6.89  & 22.99--28.86 \\
    \bottomrule
  \end{tabular}
\end{table*}

Total analyst time with DealsFlow averaged 18.85 minutes, a paired saving of 25.84 minutes (95\% CI 22.99--28.86) or 57.8\% of the manual time. Adding the 6.13 minutes of machine time gives an elapsed time of 24.98 minutes, 44.1\% below manual triage. Verification and correction took half of the assisted analyst time (50.1\%, 9.45 of 18.85 minutes). We did not break down the manual time by activity, so the source of the saving is not isolated. This was a small timing comparison, not a controlled trial. For each application one analyst performed manual triage and the other used DealsFlow, with assignments counterbalanced so that each analyst did 10 applications in each workflow and neither processed the same application twice. The paired saving therefore compares two different analysts on the same application, which counterbalancing balances across the two analysts but cannot remove. We did not measure whether analysts' conclusions differed between the two workflows, and we did not record inference cost per application.

\subsection{RQ5: Factual-support quality}
\label{sec:support}

Time saved is worth little if the report is wrong. Two reviewers independently rated the 240 distinct cited claims in the 20 reports (\S\ref{sec:analyst}); Table~\ref{tab:support} reports the results.

\begin{table}[t]
  \caption{Factual-support assessment of the 20 generated reports. Counts are after reconciliation, except agreement, which is on the reviewers' initial ratings.}
  \label{tab:support}
  \small
  \begin{tabular}{p{0.62\columnwidth}r}
    \toprule
    Measure & Result \\
    \midrule
    Reports assessed / distinct claims checked & 20 / 240 \\
    Claims fully supported by the cited source & 202 / 240 (84.2\%) \\
    Claims partly supported & 22 / 240 (9.2\%) \\
    Claims not supported & 10 / 240 (4.2\%) \\
    Cited source not accessible & 6 / 240 (2.5\%) \\
    \emph{External} or \emph{registry} labels confirmed by a cited source & 42 / 48 (87.5\%) \\
    \emph{Deck-only} or \emph{none} categories that reviewers could corroborate & 7 / 32 (21.9\%) \\
    Material issues missed in the initial reports & 9 (0.45 per report) \\
    Unsupported statements remaining after correction & 2 (0.10 per report) \\
    Corrections made & 29 (1.45 per report) \\
    Reports needing at least one correction & 14 / 20 (70.0\%) \\
    Initial reviewer agreement & 92.9\% \\
    Cohen's $\kappa$, initial ratings & 0.751 \\
    \bottomrule
  \end{tabular}
\end{table}

Most cited claims held up: 84.2\% were fully supported and 93.3\% at least partly. But the reports were not clean. Fourteen of the 20 needed at least one correction, 29 corrections were made in total, and the reviewers found 9 material issues that the initial reports did not mention. The corrections fell into four types (Table~\ref{tab:corrections}): qualifying overstated or partly supported claims (12), removing or replacing unsupported claims (8), adding omitted material issues (6) and repairing citations (3). We did not rate their severity. After correction, 2 unsupported statements remained, which equals the 10 unsupported claims less the 8 removed or replaced.

\begin{table}[t]
  \caption{Corrections made to the 20 reports, by type.}
  \label{tab:corrections}
  \small
  \begin{tabular}{p{0.62\columnwidth}r}
    \toprule
    Correction type & Count \\
    \midrule
    Qualifying overstated or partly supported claims & 12 \\
    Removing or replacing unsupported claims & 8 \\
    Adding omitted material issues & 6 \\
    Repairing inaccessible or incorrect citations & 3 \\
    \bottomrule
  \end{tabular}
\end{table}

The corroboration labels were mostly right but imperfect in both directions. Six of 48 labels of \emph{external} or \emph{registry} (12.5\%) were not confirmed by a cited source, so those categories were exempt from the ceiling on the strength of labels that the assessed citations did not substantiate. This does not show that no supporting evidence exists. Seven of 32 labels of \emph{deck\_only} or \emph{none} (21.9\%) could have been corroborated, so those categories remained subject to the ceiling although corroborating evidence existed. The two reviewers agreed on 223 of the 240 claims before reconciliation (92.9\%, $\kappa=0.751$; Appendix~\ref{app:agreement}). Taken together, the study shows shorter completion times alongside a continuing need for verification and correction. We set no quality threshold in advance and did not compare the quality of final assessments between the two workflows.

\section{Discussion and Limitations}
\label{sec:limits}

\paragraph{What the results establish.} On 142 mature investments the full pipeline's score is positively rank-correlated with the total-value multiple ($\rho=0.45$) and identifies write-offs better than chance (AUC$_0=0.71$), and the guardrails improve on an unconstrained LLM by a small but consistent margin (RQ2, RQ3). The improvement comes mostly from the sceptic pass; the corroboration ceiling adds little by itself. We did not isolate why: the lower-only rule, an extra review, or less sampling noise from combining two calls would each produce this pattern. The results do not establish screening skill (RQ1). Later successes were corroborated more often than write-offs, retrieval was not date-restricted, and the model's knowledge cutoff postdates most outcomes, so hindsight cannot be excluded as an explanation for part or all of the correlation. Discrimination is concentrated between the best outcomes and the rest; the middle groups overlap heavily. In a separate study of 20 inbound applications, total analyst time fell from 44.7 to 18.9 minutes per application (a 57.8\% reduction) even after verification and correction, and 84.2\% of 240 cited claims were fully supported by their sources (RQ4, RQ5). But 70\% of reports needed at least one correction, so the reports were not used without analyst verification. Neither result says anything about whether the system improves investment decisions.

\paragraph{Selection.} All 243 deals were funded, so we evaluate ranking within an already-screened population and cannot observe the outcomes of rejected applications. This is the VC analogue of survivorship bias \cite{brown1992survivorship,kong2026bias}. It probably makes discrimination harder to detect (the obviously weak applications are absent) and means the results say nothing about false negatives.

\paragraph{Censored outcomes.} Even after three years, many outcomes are unresolved: 26 of the 142 mature holdings are still held at cost. Marks on active holdings are the manager's valuations, not realised returns. Results could shift as the portfolio matures.

\paragraph{Residual leakage.} Retrieval is not restricted to the decision date. Web search carried no date filter, and the Companies House snapshot uses current status and officers, so a company that later dissolved can look dissolved. Search ranking may also reflect later attention \cite{kong2026bias}, and the model's knowledge cutoff (18 May 2026) postdates most outcomes, so memorisation of outcomes is not controlled \cite{lopezlira2025memorization}. We report corroboration rates by outcome class so that readers can judge the remaining risk rather than claim it is zero. The cleanest remedy is prospective evaluation on live deal flow.

\paragraph{The ceiling penalises companies with little public evidence.} Even with stage-aware values, a pre-revenue or stealth company whose factual categories rest on the deck alone is capped (at 78 if its stage is recognised as early, 70 if not), so the score partly encodes how much public evidence exists. In deployment such companies are routed to an analyst.

\paragraph{The analyst study is small and not independent.} The analyst measures rest on 20 applications and two reviewers who work with the fund manager and are not independent of the system's developers. They assessed reports that show the system's own labels and scores, so they were not blind to them, and each used DealsFlow on half of the applications whose reports they rated. Every application was triaged by both workflows but by different analysts, so the paired comparison confounds workflow with analyst; counterbalancing balances this across the two analysts but cannot remove it, and with two analysts we cannot estimate variation between analysts. The measures show whether the reports are usable and accurate, not whether the system improves investment decisions. The third reviewer, who adjudicated unresolved ratings, is an author and manages the partnership, so the adjudication was not independent either.

\paragraph{The sceptic's mechanism is not isolated.} Most of the improvement over the unconstrained LLM is already present once a second call is added. We did not compare against an equally resourced second review that may revise scores in either direction, or against averaging two independent scoring calls, and the configurations are not matched on call budget. The gain could therefore come from the lower-only permission, from the extra review, or from combining two noisy calls; the data cannot tell these apart.

\paragraph{Run-to-run variance not measured.} Scoring calls use temperature 0.1 without a fixed seed, and web evidence can differ between screens of the same deck. We did not measure how much composite scores vary across repeated runs, so the stability of an individual deal's score is unknown; all results are from a single run per deal.

\paragraph{Single fund, single model.} Results come from one fund's portfolio and investment style and one LLM. The rubric, weights and ceiling values may not transfer to other funds.

\paragraph{Untested components.} Matching, the historical adjustment, reflection and prompt-injection defences are described but not evaluated here.

\paragraph{Reproducibility.} The portfolio data, code and prompts are proprietary and are not released. We report the configuration (Appendix~\ref{app:config}), the score bands, category definitions and corroboration labels (Appendix~\ref{app:rubrics}), and all formulas and parameters, so the design can be reimplemented; the empirical results cannot be independently re-run.

\paragraph{Future work.} Prospective evaluation on live deal flow, which removes the look-ahead problem by construction; testing the matching and adjustment components against held-out outcomes; a red-team study of the injection defences; a call-budget-matched comparison of the lower-only sceptic with a reviewer that may revise scores in either direction; and repeated runs on stored evidence to measure stability.

\section{Conclusion}

DealsFlow keeps the LLM away from the composite score: the LLM reads the deck, cites evidence and rates each category, while code combines the ratings, caps them and only ever lowers them, and the sources and the scores before and after each guardrail are recorded for audit. Evaluating such a system on historical deals is easy to get wrong, because the main guardrail rewards exactly the kind of evidence that hindsight inflates. On the 142 mature investments in our 243-deal SEIS/EIS portfolio, the full pipeline's score correlates with total-value multiples ($\rho=0.45$, 95\% CI [0.32, 0.57]) and improves modestly on an unconstrained LLM, mostly through the sceptic pass. But write-offs were corroborated less often than later successes and retrieval was not date-restricted, so the results may overstate screening skill. In a separate study of 20 inbound applications, analyst time fell by 58\% even after verification and correction, and 84\% of cited claims were fully supported, but 70\% of reports still needed a correction. We hope the leakage diagnostics, more than the system itself, are useful to others evaluating LLM screening tools on historical financial data.

\begin{acks}
This work was carried out within a Knowledge Transfer Partnership between Queen's University Belfast and Sapphire Capital Partners, supported by Innovate UK.
\end{acks}

\section*{Ethics and Privacy Statement}

\paragraph{Data and privacy.} The portfolio records and application files were provided by Sapphire Capital Partners, a UK early-stage fund manager, within the Knowledge Transfer Partnership described in the acknowledgements. All figures in this paper are aggregated; no company, founder or investor is identifiable, and investor-level records were not used. Model inference ran through the Microsoft Azure API in an EU region. Under Azure's default terms, prompts and outputs are not used to train the underlying models and are not shared with other customers; they may be retained for up to 30 days for abuse monitoring unless an exemption applies. Companies House and web queries were made live and were not restricted to the investment date (\S\ref{sec:pit}). The university has confirmed that the project data may be used in this study.

\paragraph{Analyst participation.} The two analysts who timed the workflows and rated the generated reports work with the fund manager and took part as part of their normal work. No personal data about the analysts is reported.

\paragraph{Societal risks.} A screening tool that rewards public evidence can disadvantage founders with a small public footprint, which may correlate with geography, background or access to networks. The stage-aware ceiling softens this for early-stage companies but does not remove it, and we did not test for disparate impact across founder groups. Because the system can reject automatically below a threshold, such founders could be excluded without a human reading their deck; we recommend that deployments sample and review automated rejections. Our results show only a modest correlation with outcomes that may be inflated by leakage, so scores should not be read as predictions or used without human judgement. Finally, founder decks are confidential, and sending them to a hosted model carries a confidentiality risk, which we reduce by using an EU-region deployment under Azure's data terms.

\paragraph{Competing interests.} The DealsFlow codebase is owned jointly by Queen's University Belfast and Sapphire Capital Partners through the partnership. Some authors are employed by Sapphire Capital Partners, which uses DealsFlow in production and supplied the portfolio data used in this study.

\paragraph{Use of generative AI.} Generative AI tools were used for copy-editing and formatting. System design, experiments, analysis and conclusions are the authors' own.

\bibliographystyle{ACM-Reference-Format}
\bibliography{references}

\appendix

\section{Models and Configuration}
\label{app:config}

\begin{table}[!htb]
  \caption{Settings used in the experiments.}
  \small
  \begin{tabular}{p{0.34\columnwidth}p{0.58\columnwidth}}
    \toprule
    Item & Value \\
    \midrule
    Scoring model & GPT-6 Luna (\texttt{gpt-6-luna}) via Microsoft Azure, EU region; knowledge cutoff 18 May 2026 \cite{openai2026gpt6luna} \\
    Sceptic model & Same model; second call; reductions only, at most 20 points per category \\
    Sampling & Temperature 0.1 on every call; all other sampling and reasoning settings at the Azure defaults \\
    Web queries per deck & At most 10 by construction (hard cap 12) \\
    Search & \texttt{ddgs} metasearch, 5 results per query; top 2 pages crawled (\texttt{crawl4ai}), 4{,}000 characters each; results cached for 24 hours \\
    Date filter on search & None in any run \\
    Corroboration ceiling & Stage-aware: 78 (early), 70 (unknown), 60 (growth) \\
    Category weights & Vision 1.0, market 1.5, financials 1.5, founders 2.0, traction 1.5, differentiation 1.5 \\
    \bottomrule
  \end{tabular}
\end{table}

\section{Score Bands and Category Definitions}
\label{app:rubrics}

The scoring prompt uses one set of bands for all categories: 75 and above is exceptional or clearly strong; 55--74 is credible with open questions; 35--54 is weak; below 35 is poor or incoherent. There are no per-category anchors. Table~\ref{tab:rubric} gives each category's definition and what counts as corroboration.

\begin{table*}[!t]
  \caption{Category definitions, corroboration sources and ceiling status.}
  \label{tab:rubric}
  \small
  \begin{tabular}{p{0.16\textwidth}p{0.40\textwidth}p{0.30\textwidth}c}
    \toprule
    Category (weight) & What is judged & Counts as corroboration & Ceiling \\
    \midrule
    Vision \& mission (1.0) & Clarity and ambition of the vision; whether the mission follows from the problem & Judgement-based; rarely externally checkable & No \\
    Market potential (1.5) & Problem severity and realism; market-size claims sanity-checked; pricing realism & A web source on market data, competitors or customers & Yes \\
    Financials \& valuation (1.5) & Credibility of the revenue trajectory; valuation reasonable for stage and traction; quality of the raise against the plan & Companies House data, or a public funding or press source & Yes \\
    Founders (2.0) & Depth and relevance of expertise for this problem, from the deck and web research; red flags & Founder web sources or Companies House officer records & No \\
    Traction (1.5) & Real demand (revenue, pilots, signed customers, growth) rather than vanity metrics & Customer, review or press sources, or registry filings & Yes \\
    Differentiation (1.5) & Unique edge or moat; defensibility against alternatives & A web source on competitors or alternatives & Yes \\
    \bottomrule
  \end{tabular}
\end{table*}

\section{Structural Validity Checklist}
\label{app:checklist}

Table~\ref{tab:checklist} maps our evaluation onto the checklist of Kong et al.\ \cite{kong2026bias}.

\begin{table}[!htb]
  \caption{Evaluation against the structural validity framework of \cite{kong2026bias}.}
  \label{tab:checklist}
  \small
  \begin{tabular}{p{0.28\columnwidth}p{0.62\columnwidth}}
    \toprule
    Requirement & How addressed here \\
    \midrule
    Temporal sanitation (weights) &  Model cutoff (18 May 2026) disclosed; it postdates most outcomes and is not controlled; residual leakage reported (\S\ref{sec:limits}). \\
    Temporal sanitation (retrieval) & No date filter on web search; Companies House status is current; residual leakage reported (\S\ref{sec:limits}). \\
    Dynamic universe / survivorship & All 243 funded deals included, including 47 write-offs and 10 loss exits; primary set restricted to 142 deals at least 3 years old to reduce the influence of very recent investments (outcomes remain censored) (\S\ref{sec:maturity}); rejected applications unobservable (\S\ref{sec:limits}). \\
    Rationale robustness & Each category cites numbered sources;{} the stored source list (title and URL) permits source-level audit, and analysts check whether cited sources support the claims (RQ5). \\
    Epistemic calibration & Stage-3 survival estimates are not evaluated in this paper (\S\ref{sec:other}). \\
    Cost & Total analyst time including verification and correction, and machine time, per application (RQ4); inference cost not measured. \\
    \bottomrule
  \end{tabular}
\end{table}

\section{Reviewer Agreement on Claim Support}
\label{app:agreement}

Table~\ref{tab:agreement} shows the two reviewers' initial ratings of the 240 distinct claims, before reconciliation. They agreed on 223 claims (92.9\%), and unweighted Cohen's $\kappa$ is 0.751.

\begin{table}[!htb]
  \caption{Initial ratings of the 240 claims by Reviewer A (rows) and Reviewer B (columns).}
  \label{tab:agreement}
  \small
  \begin{tabular}{lrrrr}
    \toprule
    A $\backslash$ B & Supported & Partly & Unsupported & Inaccessible \\
    \midrule
    Supported        & 195 & 6  & 1 & 1 \\
    Partly supported & 4   & 15 & 2 & 0 \\
    Unsupported      & 1   & 2  & 7 & 0 \\
    Inaccessible     & 0   & 0  & 0 & 6 \\
    \bottomrule
  \end{tabular}
\end{table}

\clearpage
\end{document}